\documentclass[letterpaper, 10 pt, conference]{ieeeconf}  % Comment this line out
\IEEEoverridecommandlockouts                              % This command is only
\title{\LARGE \bf
Removal of Parameter Adjustment of Frangi Filters in Case of Coronary Angiograms
}

\author{ \parbox{3 in}{\centering Huibert Kwakernaak*
         \thanks{*Use the $\backslash$thanks command to put information here}\\
         Faculty of Electrical Engineering, Mathematics and Computer Science\\
         University of Twente\\
         7500 AE Enschede, The Netherlands\\
         {\tt\small h.kwakernaak@autsubmit.com}}
         \hspace*{ 0.5 in}
         \parbox{3 in}{ \centering Pradeep Misra**
         \thanks{**The footnote marks may be inserted manually}\\
        Department of Electrical Engineering \\
         Wright State University\\
         Dayton, OH 45435, USA\\
         {\tt\small pmisra@cs.wright.edu}}
}

\author{ Rishabh Joshi* % stops a space
\hspace*{0.01 in} Dhruv Gosain*
\thanks{* These authors have contributed equally to the project.}
}

\usepackage{graphicx}
\usepackage{subfig}
\begin{document}

\maketitle
\thispagestyle{empty}
\pagestyle{empty}

%%%%%%%%%%%%%%%%%%%%%%%%%%%%%%%%%%%%%%%%%%%%%%%%%%%%%%%%%%%%%%%%%%%%%%%%%%%%%%%%
\begin{abstract}

Frangi Filters are one of the widely used filters for enhancing vessels in medical images. Since they[2] were first proposed, the threshold of the vesselness function of Frangi Filters is to be arranged for each individual application[1]. These thresholds are changed manually for individual fluoroscope, for enhancing coronary angiogram images. Hence it is felt, there is a need of mitigating the tuning procedure of threshold values for every fluoroscope. The current paper’s approach has been devised in order to treat the coronary angiogram images uniformly, irrespective of the fluoroscopes through which they were obtained and the patient demographics for further stenosis detection. This problem to the best of our knowledge has not been addressed yet. In the approach, before feeding the image to Frangi Filters, non uniform illumination of the input image is removed using homomorphic filters and the image is enhanced using Non Subsampled Contourlet Transform (NSCT). The experiment was conducted on the data that has been accumulated from various hospitals in India and the results obtained verifies dependency removal of parameters without compromising the results obtained by Frangi filters.

\end{abstract}

%%%%%%%%%%%%%%%%%%%%%%%%%%%%%%%%%%%%%%%%%%%%%%%%%%%%%%%%%%%%%%%%%%%%%%%%%%%%%%%%
\section{INTRODUCTION}
\subsection{Frangi Filters}
Frangi Filters were introduced by Frangi et al. \cite{c2} in the year 1998, for enhancing blood vessels in the medical image. The approach assumes the enhancement process equivalent to filtering of tubular geometrical structures in the given image. The earlier methods detected vessels using a fixed scale approach\cite{c2}, which limited their approach to smaller size range. Frangi filters adopt multi-scale, thus making processing of large images more optimal.
\subsection{ Limitations }
The vesselness function as proposed by Frangi et al. mathematically represents the tubular structure present in the image taken into consideration. The thresholds in the vesselness function control the sensitivity of the proposed function to the measures of geometric ratios and the "second order structureness". Currently Frangi Filters are being used with fixed values of these thresholds\cite{c8}. These thresholds have to be varied for different fluoroscope machines since each one produces different level of noise.  
Frangi Filters also pose another problem that the vessel boundaries in the output image are narrower than the original image, since the vesselness function becomes weak at the boundaries. 
\subsection{ Contribution }
The proposed approach strives to treat all the angiogram images from different fluoroscopes through a same enhancement methodology to utilize these images for further analysis. Prior to enhancing the images using Non Subsampled Contourlet Transform (NSCT), homomorphic filters are used to remove non uniform illumination. Post enhancement, these images are filtered using Frangi Filters using same values of thresholds. 

Our approach is organized into following sections : 
\renewcommand{\theenumi}{\Alph{enumi}}
\begin{enumerate} 
\item Homomorphic Gaussian Filtration
\item Background Normalization
\item NSCT Decomposition - Enhancement - Reconstruction
\item Frangi Filters
\end{enumerate}
\section{Approach }
In order to remove the tuning of parameters (thresholds) for every fluoroscope, there is a need to reduce the effect of noise introduced by them, intensity inhomogeneity (due to organs) and enhancement of the vessels at their boundaries [2]. The following subsections explains the detailed approach (Fig. 1) adopted to achieve removal of parametric tuning.

\subsection{    Homomorphic Gaussian Filtration }

Coronary angiogram images obtained using fluoroscope suffer from non uniform intensity,  introduced due to the imperfections in the image obtaining process and the shadow of the organs as it can be seen in (
Fig. 2 (Original Image) and Fig. 3. The intensity of the same organ/artery varies in different images because of their respective positioning in that image. We employ Homomorphic Gaussian Filters as mentioned in \cite{c3} to eliminate this non uniformity, as this method being retrospective solely depends upon the information from the image and no prior information is required regarding patient demographics. This method assumes that the non-uniform intensity constitutes the low frequency portion of the image and is an additive noise. The following formula from \cite{c3} explains the removal of the additive low frequency noise from the image. 

\begin{equation}
log (u(x)) = log (v(x)) - LPF(log (v(x))) + C_N  
\end{equation}
where, $v(x)$ is the original image, LPF(.) is the low pass filter used in homomorphic filter (in our case it is Gaussian), $C_N$ is a constant added to preserve maximum intensity of the final image and $u(x)$ is the final image.
\newline
Thus, the homomorphic filters reduces the value of background noise quantified using Norm of the Hessian(S) mentioned in \cite{c2}, so reducing its value (S), leading to increase in the value of vesselness function.

\begin{figure}
\includegraphics[width=0.5\textwidth]{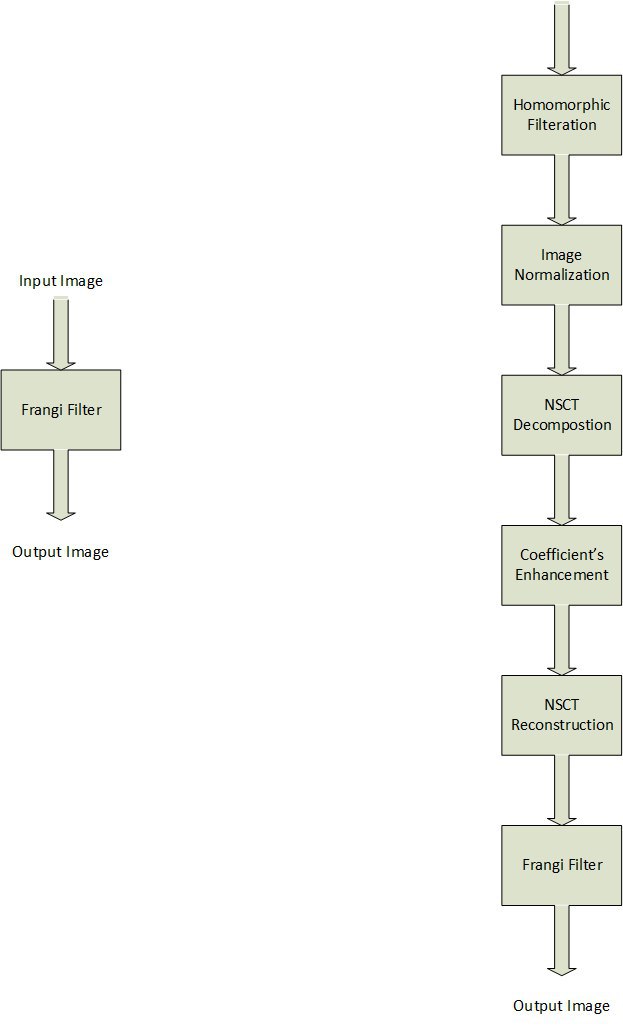}
%\subfloat[]{\includegraphics[width=0.24\textwidth]{comparishon_flow_frngi}\label{fig:f1}}
\caption{\label{fig:1A 1A}}
\end{figure} 

\begin{figure}
\includegraphics[width=0.424\textwidth]{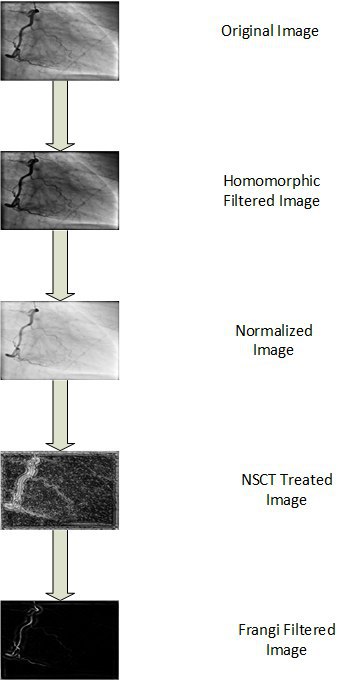}
%\subfloat[]{\includegraphics[width=0.24\textwidth]{comparishon_flow_frngi}\label{fig:f1}}
\caption{\label{fig:1A 1A}}
\end{figure} 
\subsection{    Background Normalization }

In general, the homomorphic filters while eliminating non-uniform intensity might change the overall intensity of the image. In order to normalize the image's intensity to a particular mean and variance, intensity normalization is performed [4][5]. Fig. 2 clearly exhibits the effect of normalizing the image to the desired mean and variance.
\newline \newline if img(i,j)$ >$M
\begin{equation}\label{myeq2}
N(i,j) = M_0 + \sqrt[]{\frac{VAR_0(img(i,j)-M^2)}{VAR}}  
\end{equation}
\newline otherwise
\begin{equation}\label{}
N(i,j) = M_0 - \sqrt[]{\frac{VAR_0(img(i,j)-M^2)}{VAR}}
\end{equation}
where $img(i,j)$ is the original image, $N(i,j )$ is the final image, $M$ and $VAR$ are mean and variance of the input image and $M_0$ and $VAR_0$ are mean and variance of the desired output image.
\normalsize

\subsection{ NSCT Decomposition - Enhancement - Reconstruction  }

Non-Subsampled Contourlet Transform (NSCT) method is employed to enhance the vessel boundaries. NSCT being a multi-resolution analysis method, decomposes the input image into multiscale and multi-directional sub-band images. The Non-Subsampled Pyramid (NSP) provides the multi-scale feature, whereas the shift-invariant multi-directional feature is obtained using Non-Subsampled Directional Filter Banks (NSDFB).
\newline
Since NSCT is shift invariant, each pixel at the same spatial location in the sub-bands corresponds to the pixel present at that very same spatial position in the original image, thus all the pixels in the sub-bands contribute to the original pixel. The noise pixels in the original image, which we want to get rid off, have low pixel values in all of the sub-bands whereas the weak pixels have high values in some of the sub-bands and low values in some other. Pixels at the vessel boundaries fall into this category due to high contrast between the vessel structure and background.
\newline
We employ the image denoising and enhancement technique on the high frequency sub-bands as proposed by Liu et. al. \cite{c7} to treat the sub-band images. The low frequency sub-band images are treated using contrast stretching. Finally the images are reconstructed using NSCT reconstruction method.
\newline
The enhanced image is added to the original image for high frequency emphasis. The resultant image will be similar in contents to the original image but with emphasis to the high frequency features of the image \cite{c6}. The final output of the enhancement done using NSCT can be seen in Fig. 2.
\newline
The NSCT method overcomes the limitation of weak vessel boundaries observed in case of Frangi Filters by enhancing the pixels and supressing the noise at the vessel boundaries.

\subsection{ Frangi Filters\cite{c2}}
The output image of that NSCT filter is passed using Frangi Filters. Frangi filters can be represented in terms of its vesselness function,
\begin{equation}\label{myeq4}
f(x) =   \left\{
\begin{array}{ll}
0 & $$ \lambda_2 $$ > 0 \\
e^(-\frac{(R_b)^2}{2\alpha^2}) (1 - e^(\frac{S^2}{2\beta^2})) & elsewhere
\end{array}
\right.
\end{equation}
\newline
Here $R_b$ accounts for measuring the deviation from blob like structure to plate or line like structure and S, norm of the hessian is used to quantify the signal to noise ratio (SNR) of the background pixels present in the image. Also, $ \alpha $ and $\beta $ are the threshold value of the vesselness function.
\newline The vesselness function is computed for different scales between 0.5 to 15 in increments of 0.5 and its maximum value is chosen.
\begin{equation}\label{}
(x) = max_\sigma f(x,\sigma)
\end{equation}
Here $\sigma $ represents the standard deviation for Gaussian Image Derivative. The difference in output of only Frangi Filter and our method can easily be seen in Fig. 2
\section{Experiment and Results}

Implementation of the following approach is done on Intel i7 Processor, 8 GB RAM, Windows 10, MATLAB. In this section we will discuss about the results on various angiography sources. We in total ran this procedure on 4 sets consisting of nearly 20 images from different machines collected from several hospitals. 
\subsection{Results}
Fig 3 shows two original coronary artery images. Fig 4 (a) and (b) contains the ground truth images observed with the help of interventional cardiologist for the original images shown in Fig 3. We applied fig 3 (a) and (b) as input to Frangi Filter with value of $ \alpha $ and $\beta $ as 0.5 and 15 respectively. The output of Frangi Filters can be seen in Fig 5 (a) and (b). While applying our procedure on the original images we can see a huge improvement in artery segmentation(Fig 6 (a) and (b)). The area under ROC curve increased from 0.0836 of only Frangi to 0.2731 of our technique for image 3(b) (Fig 8) and from 0.0544 to 0.1751 for image 3(a) (Fig 7). This showed that even without changing the values of the threshold for different machines, we are able to get the desired results, thus removing the need of parameter adjustment.

\subsection{Improvements in output image}
In this paper we take area under ROC curve as paradigm for comparing our procedure results with only Frangi Filter's result. Here after running the procedure on several images we observed that for every image we are improving the output of Frangi Filters. In Fig 6(a,b) the area under ROC curve increased from 0.0836 of only frangi to 0.2731 of our technique for image 3(b) (Fig 8) and from 0.0544 to 0.1751 for image 3(a) (Fig 7). Thus our procedure is not only removing the tuning process of threshold values but also improving the quality of output of Frangi Filters.

\begin{figure}
\subfloat[]{\includegraphics[width=0.24\textwidth]{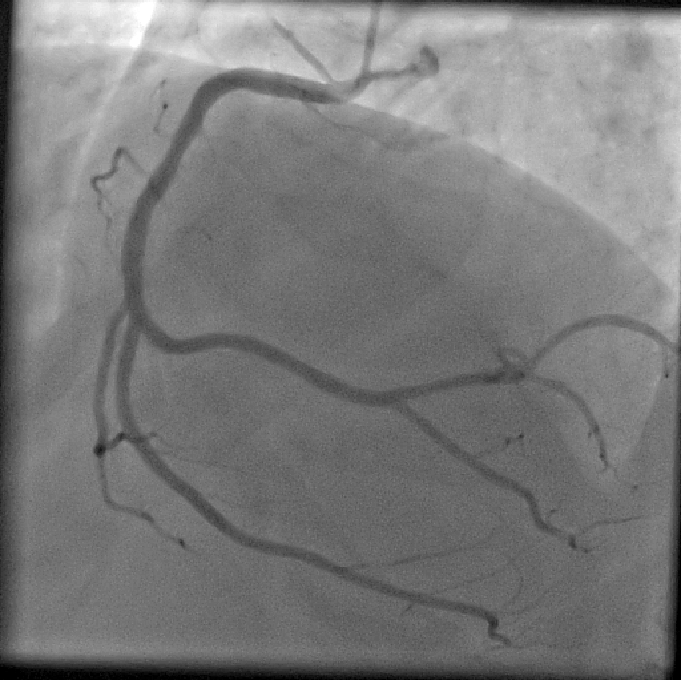}\label{fig:f1}}
\subfloat[]{\includegraphics[width=0.24\textwidth]{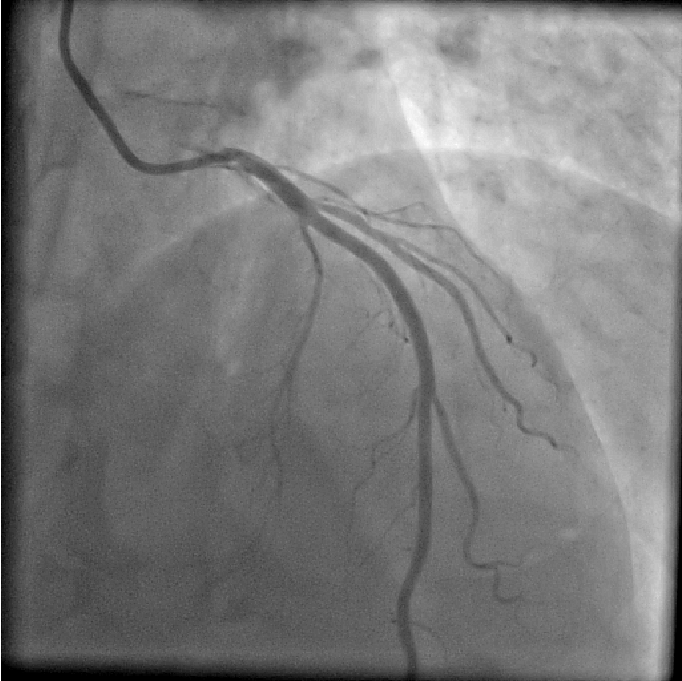}\label{fig:f2}}
\caption{\label{fig:1A 1A}}
\end{figure}
\begin{figure}
\subfloat[]{\includegraphics[width=0.24\textwidth]{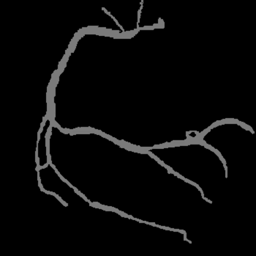}\label{fig:f1}}
\subfloat[]{\includegraphics[width=0.24\textwidth]{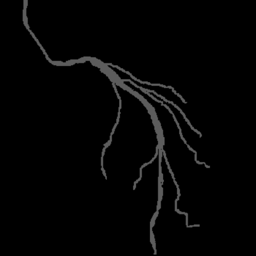}\label{fig:f2}}
\caption{\label{fig:1A 1A}Ground truth images}
\end{figure}
\begin{figure}
\subfloat[]{\includegraphics[width=0.24\textwidth]{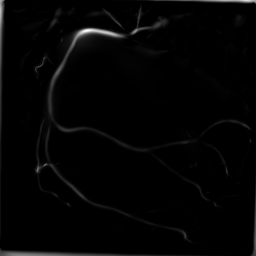}\label{fig:f1}}
\subfloat[]{\includegraphics[width=0.24\textwidth]{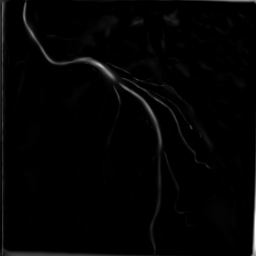}\label{fig:f2}}
\caption{\label{fig:1A 1A} Output of frangi filters }
\end{figure}
\begin{figure}
\subfloat[]{\includegraphics[width=0.24\textwidth]{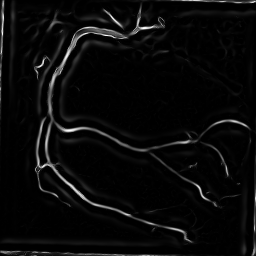}\label{fig:f1}}
\subfloat[]{\includegraphics[width=0.24\textwidth]{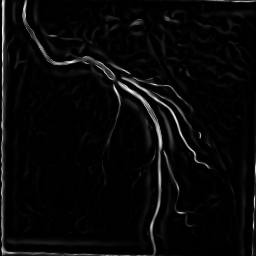}\label{fig:f2}}
\caption{\label{fig:1A 1A} Output of our approach }
\end{figure}
\begin{figure}
\includegraphics[width=0.55\textwidth]{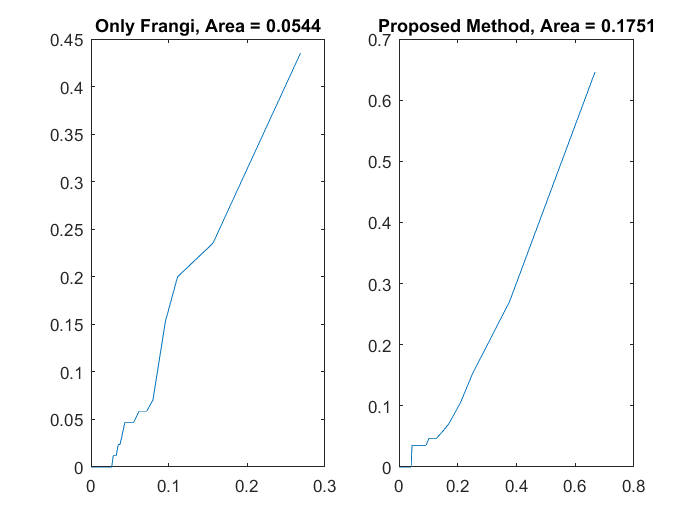}
%\subfloat[]{\includegraphics[width=0.24\textwidth]{final_test_image_2}\label{fig:f1}}
%\subfloat[]{\includegraphics[width=0.24\textwidth]{final_image}\label{fig:f2}}
\caption{\label{fig:1A 1A} ROC curve values for Fig 3(a) }
\end{figure}
\begin{figure}
\includegraphics[width=0.52\textwidth]{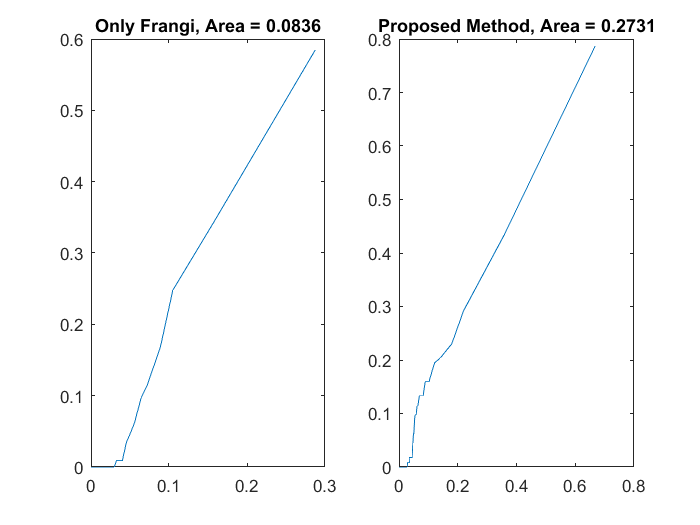}
%\subfloat[]{\includegraphics[width=0.24\textwidth]{final_test_image_2}\label{fig:f1}}
%\subfloat[]{\includegraphics[width=0.24\textwidth]{final_image}\label{fig:f2}}
\caption{\label{fig:1A 1A} ROC curve values for Fig 3(b) }
\end{figure}
\subsection{Limitation}
While running the procedure on several images, in some cases(fig 9) by changing the value of $ \alpha $ and $\beta $ further improvement in area under ROC curve is observed, this shows that there is moderate contribution of parameters in the output in this case. Though even without changing the values of parameters it is giving satisfactory results but we can still achieve a better result by adjusting parameter values. This led us to our future work of reaching an absolute optimal value of frangi filter and obsoleting the interference of parameters in vesselness function output.    

\begin{figure}
\subfloat[]{\includegraphics[width=0.24\textwidth]{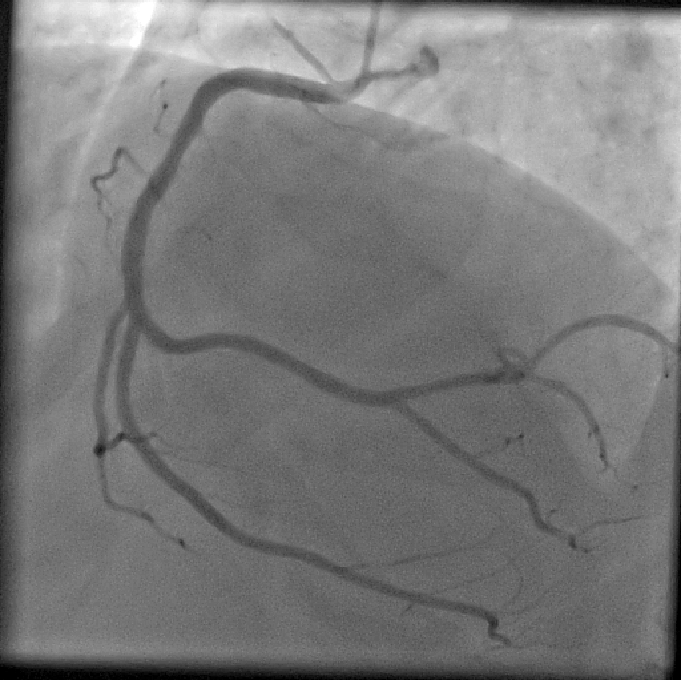}\label{fig:f1}}
\subfloat[]{\includegraphics[width=0.24\textwidth]{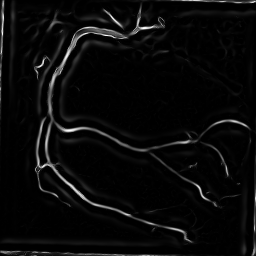}\label{fig:f1}}
%\subfloat[]{\includegraphics[width=0.24\textwidth]{original_image}\label{fig:f2}}
\caption{\label{fig:1A 1A}}
\end{figure}

\section{Discussion}
During the course of our research, it has been observed that Gaussian Homomorphic Filters provides better visual results than Butterworth Homomorphic Filters. It is also observed that there is a direct relationship between the area under the ROC curve and the standard deviation of Gaussian homomorphic filters, since the standard deviation quantifies the bandwidth of the low frequency noise which is removed from the source image. 

\section{CONCLUSION}

Through our research, we have made an attempt to remove the process of tuning of threshold values for every fluoroscope. In order to achieve this, the discussed approach has been experimented with the above mentioned dataset, whose results are justified using the vesselness function. The homomorphic filtration process has reduced the background noise and non uniform illumination, thus lowering the value of Norm of the Hessian(S) in the vesselness function, which in turn decreased the contribution of background noise. Additionally the NSCT process enhanced the weaker boundaries of the arteries, leading to an increase in the volume of the artery and $R_b$ (used to measure deviation from blob like structure) value (\ref{myeq4}) which raised the value of vesselness function. The values of $R_b$ and S (\ref{myeq4}) have been increased and decreased respectively to such extent that they dominate over the threshold values, thus removing the need to tune them for every individual fluoroscope. However, while running the procedure on several images, it is observed in some cases that by changing the value of parameters (which were fixed initially) area under ROC curve has also increased, which shows that the parameters values are contributing moderately in these cases which gives us the scope to our future work of making values of parameters obsolete in vesselness function value determination.

\section{FUTURE WORK}
We want to further optimize our approach so as to optimal results for coronary artery segmentation regardless of any value of $\alpha$ and $\beta$(threshold values of Frangi Filter) such that with any value of $ \alpha$ and $\beta$ we get a constant result.
\addtolength{\textheight}{-12cm}   % This command serves to balance the column lengths

%%%%%%%%%%%%%%%%%%%%%%%%%%%%%%%%%%%%%%%%%%%%%%%%%%%%%%%%%%%%%%%%%%%%%%%%%%%%%%%%

%%%%%%%%%%%%%%%%%%%%%%%%%%%%%%%%%%%%%%%%%%%%%%%%%%%%%%%%%%%%%%%%%%%%%%%%%%%%%%%%

%%%%%%%%%%%%%%%%%%%%%%%%%%%%%%%%%%%%%%%%%%%%%%%%%%%%%%%%%%%%%%%%%%%%%%%%%%%%%%%%
%\section*{APPENDIX}


%\section*{ACKNOWLEDGMENT}

%The preferred spelling of the word ÒacknowledgmentÓ in America is without an ÒeÓ after the ÒgÓ. Avoid the stilted expression, ÒOne of us (R. B. G.) thanks . . .Ó  Instead, try ÒR. B. G. thanksÓ. Put sponsor acknowledgments in the unnumbered footnote on the first page.

%%%%%%%%%%%%%%%%%%%%%%%%%%%%%%%%%%%%%%%%%%%%%%%%%%%%%%%%%%%%%%%%%%%%%%%%%%%%%%%%

%References are important to the reader; therefore, each citation must be complete and correct. If at all possible, references should be commonly available publications.

\end{document}